\documentclass[10pt,journal,compsoc,twocolumn ]{IEEEtran}

\usepackage{times}
\usepackage{epsfig}
\usepackage{graphicx}
\usepackage{amsmath}
\usepackage{amssymb}
\usepackage{stfloats}
\usepackage{multirow}
\usepackage{bbm}
\usepackage{rotating}
\usepackage{array}
\usepackage{color}
\usepackage{threeparttable}

\usepackage{epstopdf}
\usepackage{pifont}
\usepackage{subfigure}
\usepackage{bm}
\usepackage{url}
\usepackage[table]{xcolor}
\usepackage{color}
\usepackage{booktabs}
\usepackage[linesnumbered,ruled,vlined]{algorithm2e}

\begin{document}

\title{Marginal Debiased Network for Fair Visual Recognition}

\author{Mei Wang, Weihong Deng, Jiani Hu, Sen Su
\thanks{Mei Wang is with School of Artificial Intelligence, Beijing Normal University, Beijing, 100875, China. E-mail: wangmei1@bnu.edu.cn.}
\thanks{Weihong Deng, Jiani Hu and Sen Su are with School of Artificial Intelligence, Beijing University of Posts and Telecommunications, Beijing, 100876, China. E-mail: \{whdeng,jnhu,susen\}@bupt.edu.cn. (Corresponding Author: Jiani Hu)}}

\markboth{Journal of \LaTeX\ Class Files,~Vol.~18, No.~9, September~2020}%
{Marginal Debiased Network for Fair Visual Recognition}

\maketitle

\begin{abstract}

Deep neural networks (DNNs) are often prone to learn the spurious correlations between target classes and bias attributes, like gender and race, inherent in a major portion of training data (bias-aligned samples), thus showing unfair behavior and arising controversy in the modern pluralistic and egalitarian society. In this paper, we propose a novel marginal debiased network (MDN) to learn debiased representations. More specifically, a marginal softmax loss (MSL) is designed by introducing the idea of margin penalty into the fairness problem, which assigns a larger margin for bias-conflicting samples (data without spurious correlations) than for bias-aligned ones, so as to deemphasize the spurious correlations and improve generalization on unbiased test criteria. To determine the margins, our MDN is optimized through a meta learning framework. We propose a meta equalized loss (MEL) to perceive the model fairness, and adaptively update the margin parameters by meta-optimization which requires the trained model guided by the optimal margins should minimize MEL computed on an unbiased meta-validation set. Extensive experiments on BiasedMNIST, Corrupted CIFAR-10, CelebA and UTK-Face datasets demonstrate that our MDN can achieve a remarkable performance on under-represented samples and obtain superior debiased results against the previous approaches.

\end{abstract}


\section{Introduction}

Recent deep neural networks (DNNs) \cite{he2016deep} have shown remarkable performance and have been deployed in some high-stake applications, including criminal justice and loan approval. However, many concerns have arisen that they often show discrimination towards bias attributes like gender and race, leading to serious societal side effects \cite{wang2021meta}. For example, COMPAS used by the United States judiciary is prone to associate African-American offenders with a high likelihood of recidivism compared with Caucasian \cite{angwin2016machine}.

\begin{figure}
\centering
\includegraphics[width=8.5cm]{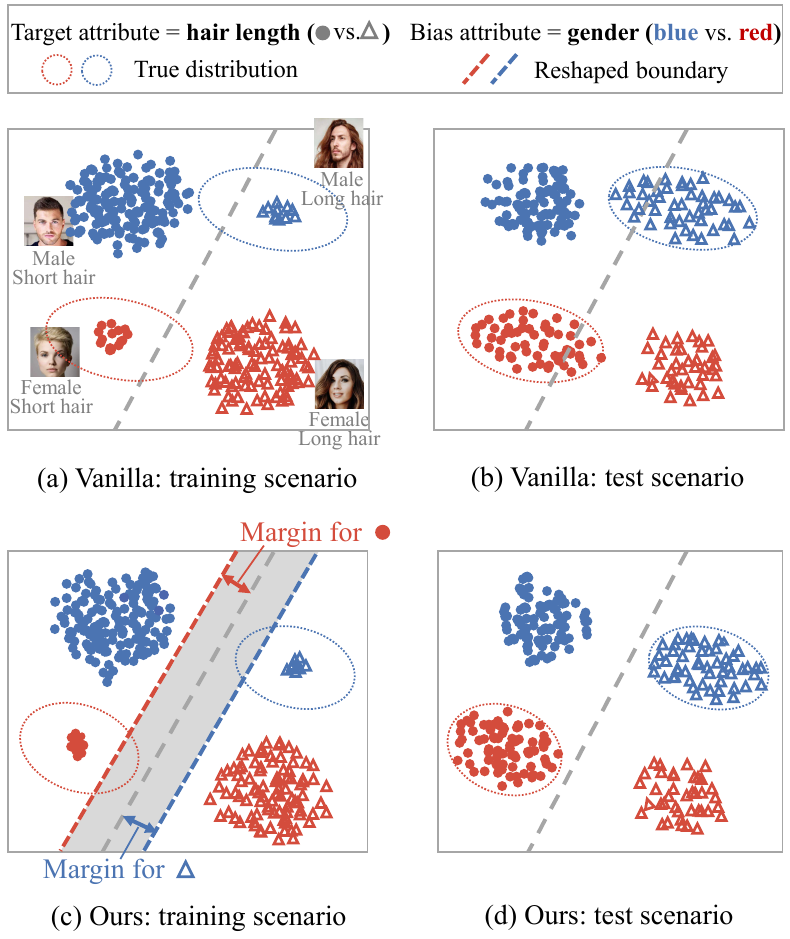}
\caption{The vanilla model learns the spurious correlations from bias-aligned samples (short-haired men) while ignoring bias-conflicting samples (short-haired women), leading to a biased problem (upper). Our MDN introduces a margin penalty and leverages meta learning to assign stronger constrains for bias-conflicting samples, which achieves fair performance when evaluated on unbiased datasets (lower).}
\label{overview} 
\end{figure}

A major driver of model bias is the spurious correlations between target classes and bias attributes that appear in a majority of training data, referred to as \emph{dataset bias} \cite{torralba2011unbiased}. In the case of hair length classification shown in Fig. \ref{overview}(a) and 1(b), for example, hair length is highly correlated to gender attribute, i.e., most people with short hair (i.e., target class) are male (i.e., bias attribute) in a dataset. When DNNs focus more on male images with short hair while ignoring few female images with short hair, unintended correlation contained in the majority of data would mislead models to take gender attribute as a cue for classification,
leading to high performance for short-haired males but worst-case errors for short-haired females.
Besides, since female training images with short hair are rare, they often fail to describe the true distribution with large intra-class variation, which makes models unable to generalize well on new test samples and thereby leads to inaccurate inferences as well. Throughout the paper, we call data with such spurious correlations and constituting a majority of training data \emph{bias-aligned samples} (e.g., male with short hair), and the other \emph{bias-conflicting samples} (e.g., female with short hair).

How to get rid of the negative effect of the misleading correlations? One intuitive solution is to enforce models to emphasize bias-conflicting samples and improve their feature representations when training. Existing methods either overweight (oversample) them \cite{nam2020learning} or augment them by generative models \cite{kim2021biaswap}. However, the former cannot address intra-class variation and results in limited improvement of generalization capability, while the latter often suffers from model collapse when training generative models.

In this paper, we focus on the spurious correlations and the generalization gap of different subgroups, and aim to mitigate bias in visual recognition from the perspective of margin penalty. To this end, we propose a marginal debiased network (MDN). Inspired by large margin classification \cite{liu2016large}, we design a new marginal softmax loss (MSL) which imposes different margins for bias-conflicting and bias-aligned samples. To reduce the negative effects of spurious correlations, our MDN prefers to regularize bias-conflicting samples more strongly than bias-aligned ones so that models would pay more attention to the former and are prevented from learning unintended decision rules from the latter. Furthermore, previous work \cite{cao2019learning} has demonstrated that margin penalty leads to lower generalization error. Although few bias-conflicting samples are not representative enough to describe the true distribution, margin can reshape their decision boundary and strongly squeeze their intra-class variations to improve generalization on unbiased test criteria.

To determine the optimal margins, we further incorporate a meta learning framework in our MDN and learn margin parameters adaptively to trade off bias-conflicting and bias-aligned samples in each class. Specifically, a meta equalized loss (MEL) is performed on a unbiased meta-validation set and designed to help the model to distinguish fairness from bias. We jointly optimize the network and margin parameters: the \emph{inner-level} optimization updates the network supervised by MSL over training dataset for better representation learning under the reshaped boundary, while the \emph{outer-level} optimization tunes margins supervised by MEL over meta-validation set for making the network unbiased. In this way, the objectives of fairness and accuracy are split into two separate optimizations in different loops, which enables to reduce bias while guaranteeing the model performance.

Our contributions can be summarized into three aspects.

1) We propose a novel marginal debiased network to mitigate bias. A margin penalty is introduced to emphasize bias-conflicting samples, which reduces the negative effect of target-bias correlation and improves the generalization ability.

2) We develop a meta learning framework to automatically learn the optimal margins based on the guidance of meta equalized loss and backward-on-backward automatic differentiation. 

3) We conduct extensive experiments on BiasedMNIST, Corrupted CIFAR-10, CelebA and UTK-Face datasets, and the results demonstrate the proposed method shows more balanced performance on different samples regardless of bias attribute.

\section{Related work}

\subsection{Algorithmic fairness}

Bias mitigation techniques have attracted increased attention in recent times. A common practice for mitigating bias is re-weighting or re-sampling \cite{nam2020learning,ahn2022mitigating}. 
For example, LfF \cite{nam2020learning} showed that bias-aligned samples are learned faster than bias-conflicting ones, and cast the relative difficulty of each sample into a weight. Ahn et al. \cite{ahn2022mitigating} claimed that bias-conflicting samples have a higher gradient norm and proposed a gradient-norm-based dataset oversampling method. Another effective way to achieve fairness is by using data augmentation to balance the data distribution \cite{kim2021biaswap,fang2024fairness}. For example, BiaSwap \cite{kim2021biaswap} generated more bias-conflicting images via style-transfer for training. Ramaswamy et al. \cite{ramaswamy2021fair} proposed to perturb vectors in the GAN latent space to generate training data that is balanced for each bias attribute. Additionally, there are several other work to resolve dataset bias by confounding or disentangling bias information. Inspired by InfoGAN, LnL \cite{kim2019learning} proposed to minimize the mutual information between feature embedding and bias, making the features contain less bias information. In practice, it played the minimax game between feature extractor $f$ and bias prediction network $h$. $h$ is trained to predict the bias, while $f$ is trained to make the bias prediction difficult.
EnD \cite{tartaglione2021end} devised a regularization term with a triplet loss formulation to minimize the entanglement of bias features. Zhang et al. \cite{zhang2022fair} designed controllable shortcut features to replace bias features such that the negative effect of bias can be removed from target features. 
Orthogonally to these work above, Barbano et al. \cite{barbano2022unbiased} and we both introduce margin penalty into debiasing, but our algorithm is different from theirs. Barbano et al. \cite{barbano2022unbiased} proposed 
a margin-based $\epsilon$-SupInfoNCE loss which defines the margin as the minimal distance between positive and negative samples in contrastive learning and aims at increasing it to achieve a better separation between target classes, whereas we define the margin as the distance of the boundary shift and add it to logits to trade off bias-conflicting and bias-aligned samples. Moreover, the margin in $\epsilon$-SupInfoNCE is a predefined hyper-parameter, whereas we learn the margin adaptively via a meta-learning framework. 

\begin{figure*}
\centering
\includegraphics[width=17cm]{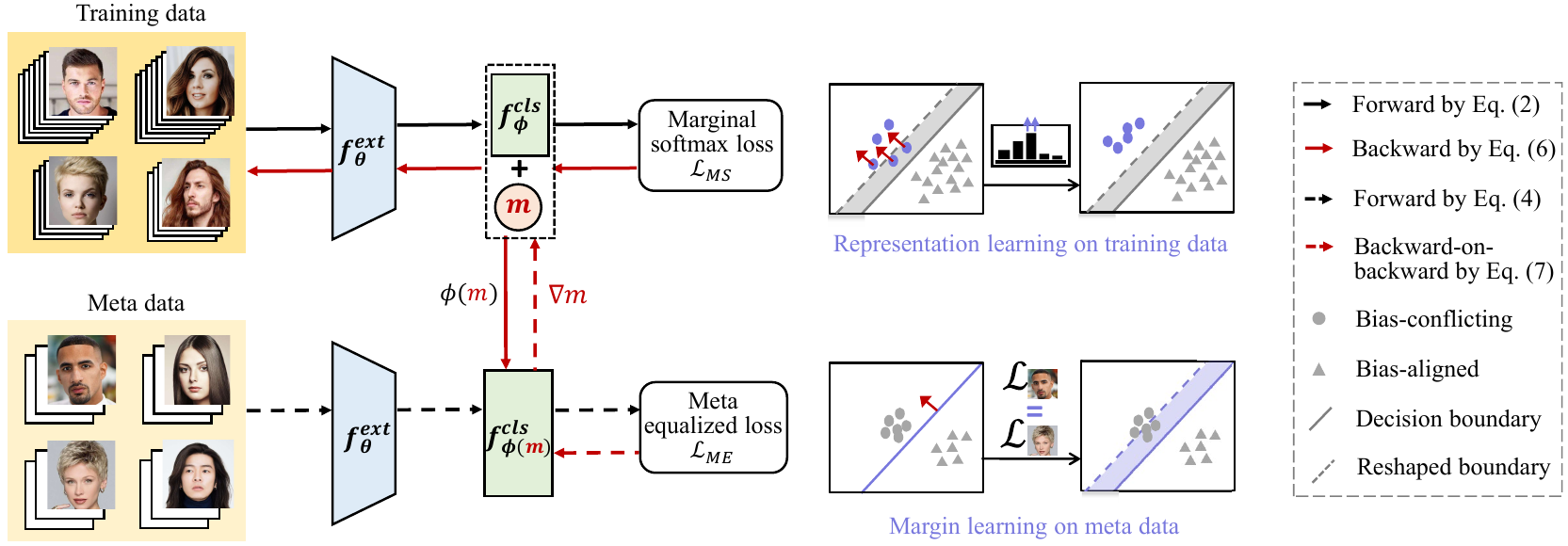}
\caption{Illustration of MDN. First, we perform representation learning on training data, which contains two main steps. i) Forward process to compute the marginal softmax loss in Eq. (\ref{source}); ii) Backward process to update the network parameters $\theta$ and $\phi \left ( m \right ) $ using Eq. (\ref{6}), where $m$ is variable used for parameterizing $\phi$. Second, we perform margin learning on meta-validation data, including two steps. iii) Forward process to compute the meta equalized loss according to Eq. (\ref{MEL}); iv) Backward-on-backward to update the margin parameter $m$ using Eq. (\ref{opt_margin}). }
\label{architecture} 
\end{figure*}

\subsection{Neural classifier}

Recently, neural classifier has been drawing extensive attention. 1) Softmax classiﬁers learn weight vector and bias term for each class, and have become the de facto regime in computer vision. Our MDN is designed based on Softmax classiﬁer, which helps to achieve fair performance. 2) Centroid-/prototype-based classifiers are nonparametric, classifying data samples based on their proximity to representative prototypes or means of classes. NCM \cite{guerriero2018deepncm} investigated the idea of bringing nearest centroids into CNNs. To discover the latent data structure and handle intra-class variations, DNC \cite{wangvisual} further estimated sub-centroids using within-class clustering. Zhou et al. \cite{zhou2022rethinking} proposed a non-learnable prototype-based semantic segmentation method that represented each class with multiple prototypes and utilized pixel-prototype contrastive learning to achieve better representation. 3) Generative classifiers learn the class densities $p(x|c)$ in contrast to discriminative classifiers which learn the class boundaries $p(c|x)$, where $x$ and $c$ represent data and labels. GMMSeg \cite{liang2022gmmseg} built Gaussian mixture models via expectation-maximization and learned generative classification with end-to-end discriminative representation in a compact and collaborative manner.

\subsection{Large margin classification}

The hinge loss is often used to obtain a “max-margin” classifier, most notably in SVMs.  With the development of deep learning, there is also a line of work \cite{wang2018cosface} focusing on exploring the benefits of encouraging large margin in the context of deep networks. Elsayed et al. \cite{elsayed2018large} designed a novel loss function based on a first-order approximation of the margin to enhance the model robustness to corrupted labels and adversarial perturbations. 
CosFace \cite{wang2018cosface} further designed an additive cosine margin to minimize the intra-class variation and enlarge the inter-class distance. 
Kim et al. \cite{kim2022adaface} proposed an adaptive margin function by approximating the image quality with feature norms, which can be utilized to emphasize samples of different difficulties and improve the recognition performance on low-quality data. Wang \cite{wang2018deep} presented a novel deep ranking model with feature learning and fusion by learning a large adaptive margin between the intra-class distance and inter-class distance. 
Our approach is most related to CosFace \cite{wang2018cosface} and Adaptiveface \cite{liu2019adaptiveface} since we both introduce margins into Softmax classifier. However, our objective and algorithm are different from those. 1) CosFace and Adaptiveface aim to learn discriminative features, whereas we focus on achieving fairness. 2) CosFace and Adaptiveface insert the margins between two classes on a hypersphere manifold, while the margins in our MSL indicate the distance of the boundary shift, i.e., the space between the decision boundaries of training and testing. 3) The margins are constants in CosFace, and are adaptively adjusted by a regularization term in Adaptiveface. Different from them, we encourage larger margins for bias-conflicting samples and incorporate a meta learning framework to learn margins adaptively.

\subsection{Meta learning}

Meta-learning, or learning-to-learn, has recently emerged as an important direction for developing algorithms in various deep-based applications. 
An inner-loop phase optimizes the model parameters using training data and an outer-loop phase updates the parameters on meta-validation data such that the model can generalize well to the validation data. MAML \cite{finn2017model} used a meta-learner to learn a good initialization for fast adapting a model to a new task. To further simplify MAML, Reptile \cite{nichol2018first} removed re-initialization for each task and only used first-order gradient information. 
Sun et al. \cite{sun2019meta} proposed a novel meta-transfer learning to learn to transfer large-scale pre-trained DNN weights for solving few-shot learning task. RMAML \cite{tabealhojeh2023rmaml} developed a geometry aware framework that generalizes the bi-level optimization problem to the Riemannian setting. 
Li et al. \cite{li2018learning} simulated train/test domain shift via the meta-optimization objective to achieve good generalization ability to novel domains. 
Martins et al. \cite{martins2023meta} proposed an online tuning of the uncertainty sampling threshold using a meta-learning approach.
MW-Net \cite{shu2019meta} learned an explicit weighting function for adaptively assigning weights on training samples to address class imbalance and label noise. In this paper, our objective is to learn an unbiased model using biased data. To achieve this, we employ a biased dataset as the training set and utilize an unbiased dataset as the meta-validation set. We take consideration of fairness, and concurrently update both the model and margin parameters through the nested optimization which ensures better generalization on unbiased test sets. 

\section{Methodology}


Let $\mathcal{D}_{train}=\{x_i,y_i,b_i\}_{i=1}^N$ be the training set, where $x_i$ is the input data, $y_i\in \{0,1,...,C\}$ denotes the target label, $b_i\in \{0,1,...,B\}$ represents the label of bias attribute (e.g., gender, age, race), $C$ and $B$ are the number of target and bias classes, and $N$ is the number of training data. $C$ can be different from $B$ but here, for clarity and without losing generality, we consider the case of $C=B$. We say that $\mathcal{D}_{train}$ is biased, i.e., the bias attribute is spuriously correlated with the target label (as an example male with short hair above). 
Given $\mathcal{D}_{train}$, our goal is to learn a classification model that makes indiscriminative predictions across the biases on unbiased test criteria. We denote a deep classification model $\Psi= f_\phi ^{cls}\circ f_\theta  ^{ext}$  with the feature extractor $f_\theta^{ext}$ and the linear classifier $f_\phi ^{cls}$, where $\theta$ and $\phi$ are the parameters of $f_\theta  ^{ext}$ and $f_\phi ^{cls}$, respectively, and $\circ$ is the function composition operator.

\subsection{Marginal softmax loss} \label{margin softmax}

The classification model is usually trained with traditional cross-entropy loss, formulated as follows,
\begin{equation}
\mathcal{L}_{CE}^i= - \log \frac{e^{\eta_{y_i}} }{\sum_k e^{\eta_k }} , \label{source}
\end{equation}
where $\mathcal{L}_{CE}^i$ denotes the loss computed on $x_i$ and $\eta_{k}$ is the logit of $x_i$ belonging to target class $k$. However, in such scenario, the learned model is easily biased and fails on the test data distribution
with the shifted correlation between target and bias attributes, i.e., $P_{train}\left ( Y,B \right ) \neq P_{test}\left ( Y,B \right )$. 
Hence our method tackles the bias problem by introducing a margin penalty and reshaping the training boundary to accommodate the distribution shift between training and testing. 

\begin{figure*}
\centering
\subfigure[Softmax: $b=0 \&1$]{
\label{margin_soft} 
\includegraphics[height=3cm]{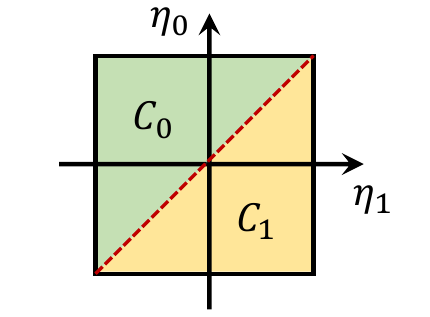}}
\hspace{0.1cm}
\subfigure[CosFace: $b=0 \& 1$]{
\label{margin_cos} 
\includegraphics[height=3cm]{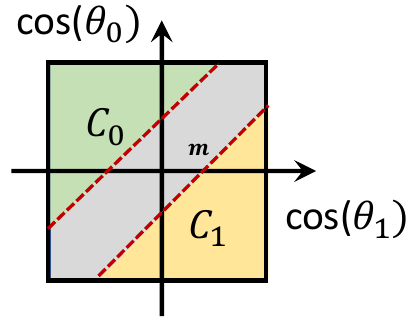}}
\hspace{0.1cm}
\subfigure[MSL: $b=0$]{
\label{margin_b0} 
\includegraphics[height=3cm]{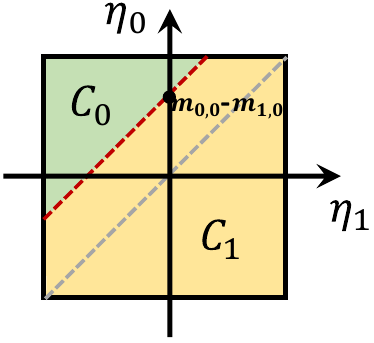}}
\hspace{0.1cm}
\subfigure[MSL: $b=1$]{
\label{margin_b1} 
\includegraphics[height=3cm]{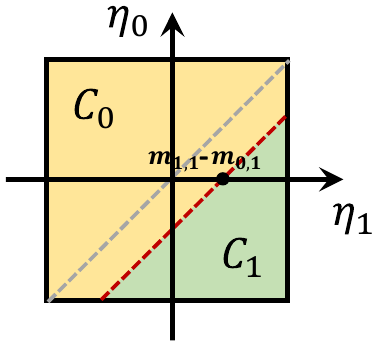}}
\caption{The comparison of decision boundaries for different loss functions in the binary-classes scenarios. Red dashed line represents the decision boundary when training, and the gray one is the decision boundary when testing. (We assume that the images from target class 0 are bias-conflicting samples when $b=0$ and the opposite when $b=1$.)}
\label{margin} 
\end{figure*}

We give a simple example to explain our idea. Consider the binary classification ($y\in\{0,1\}$ and $b\in\{0,1\}$) and we have a sample $x$ with $b=0$. 
When $b=0$, let the images with $y=0$ as bias-conflicting samples and the ones with $y=1$ as bias-aligned samples.
As shown in Fig. \ref{margin_soft}, the decision boundary of traditional cross-entropy loss is $\eta_{0}-\eta_{1}=0$, which requires $\eta_{0}>\eta_{1}$ to correctly classify $x$ from target class 0 and similarly from target class 1. 
However, we want to make the classification rigorous for bias-conflicting samples and relaxed for bias-aligned ones in order to shift the model focus from the latter to the former and mitigate the negative effect of dataset bias. To this end, we introduce the margin via a parameter $\{m_{y,b}\}_{y,b=0}^{C,B}$ for each (\emph{target, bias}) paired group to quantitatively control the decision boundary, and reshape it into $\left ( \eta_{0}-m_{0,0} \right ) -\left ( \eta_{1}-m_{1,0} \right ) =0$ for $x$ with $b=0$, where $m_{0,0}>m_{1,0}$ means a larger margin parameter is assigned for bias-conflicting samples. Thus, we force $\eta_{0}-\left ( m_{0,0}- m_{1,0}\right ) >\eta_{1}$ to correctly classify bias-conflicting samples from class 0, as shown in Fig. \ref{margin_b0}. It is essentially making the decision more stringent than previous, because $\eta_{0}-\left ( m_{0,0}- m_{1,0}\right )$ is lower than $\eta_{0}$. Similarly, if we require $\eta_{1}-\left ( m_{1,0}- m_{0,0}\right ) >\eta_{0}$ to correctly classify bias-aligned samples from class 1, the constraint would be more relaxed since $\eta_{1}-\left ( m_{1,0}- m_{0,0}\right )$ is higher than $\eta_{1}$.

Different from open-set face recognition where the margins \cite{wang2018cosface} are inserted between two classes for feature discrimination, \emph{the margin here is designed for fairness in closed-set classification and indicates the distance of the boundary shift}, i.e, the space between the decision boundaries of training and testing. As shown in Fig. \ref{margin_b0}, bias-conflicting samples are applied a positive boundary shift while bias-aligned samples receive a negative boundary shift. The reshaped decision boundary in training would strongly squeeze the intra-class variations for bias-conflicting samples, leading to improved generalization ability when testing.

Formally, we define the marginal softmax loss (MSL) as:
\begin{equation}
\mathcal{L}_{MS}^i= - \log \frac{e^{\left ( \eta_{y_i}-m_{y_i,b_i}\right ) } }{\sum_k e^{\left ( \eta_k-m_{k,b_i} \right )  }} , \label{source}
\end{equation}
where $m_{y_i,b_i}$ is the margin parameter for bias class $b_i$ and target class $y_i$, which can trade off bias-conflicting and bias-aligned samples in each target class and thus debias the model.

\subsection{Meta margin learning}

To set the optimal margin for fairness, we take $m$ as hyper-parameter associated with $\phi$ and design a meta-learning framework to jointly optimize the network parameters $\{\theta,\phi\}$ and hyper-parameter $m$, as illustrated in Fig. \ref{architecture}.

\textbf{Network parameter optimization}. Given the learned margin parameter $m$, we optimize $\{\theta,\phi\}$ supervised by MSL over training dataset under the reshaped boundary, so as to make the model to correctly classify samples from different target classes. The optimization can be formulated as follows:
\begin{equation}
\left ( \theta^* ,\phi^{*}\left ( m \right )   \right ) = \arg\underset{\left ( \theta ,\phi  \right ) }{{\min}}\ \frac{1}{N}\sum_{i=1}^N \mathcal{L}_{MS}^i \left ( \theta ,\phi, m \right ) .
\end{equation}

\textbf{Margin parameter optimization}. Given the learned model parameters $\{ \theta ,\phi\left ( m \right ) \}$, we aim to update $m$ for MSL such that the model can well trade off bias-conflicting and bias-aligned samples and perform fairly. 
To achieve this, we should be able to perceive the bias degree of the learned model at first. However, a wrong perception of bias would be obtained based on the biased training set. Therefore, we introduce an unbiased meta-validation set $\mathcal{D}_{meta}=\{x_i,y_i,b_i\}_{i=1}^M$ to simulate the shifted correlation when deployed in real-world scenarios and propose to utilize a meta equalized loss $\mathcal{L}_{ME}$ computed on $\mathcal{D}_{meta}$ to evaluate bias of $\{ \theta ,\phi\left ( m \right ) \}$, where $M$ is the number of meta-validation data. We define the meta equalized loss as follows,
\begin{equation}
\mathcal{L}_{ME} = \sum_{y\in \mathcal{Y}} \left | \frac{1}{\left | \mathcal{S}_{C}^y \right | } \sum_{i\in \mathcal{S}_{C}^y} \mathcal{L}_{CE}^i-\frac{1}{\left | \mathcal{S}_{A}^y \right | } \sum_{j\in \mathcal{S}_{A}^y} \mathcal{L}_{CE}^j \right | , \label{MEL}
\end{equation}
where $\mathcal{S}_{C}^y$ represents the image set containing bias-conflicting data in the $y$-th target class, and similarly, $\mathcal{S}_{A}^y$ is the image set containing bias-aligned data in the $y$-th target class. Bias-conflicting data refers to the (\emph{target, bias}) paired group with fewer samples, whereas bias-aligned data is the opposite. Given the labels of target classes and bias attributes, we can easily construct $\mathcal{S}_{C}^y$ and $\mathcal{S}_{A}^y$ for MEL. Therefore, $\mathcal{L}_{ME}$ represents the performance skewness of $\{\theta,\phi \left ( m \right )\}$ across different samples. Note that we herein assume that $m$ is directly related to the linear classifier $f_\phi ^{cls}$ via $\phi\left ( m \right )$. Therefore, we learn $m^*$ based on this assumption: we tune $m$ to make the linear classifier $\phi\left ( m \right )$ generalize well, or in other words, minimize $\mathcal{L}_{ME}$, when evaluated on $\mathcal{D}_{meta}$. We formulate a meta equalized loss minimization
problem with respect to $m$ as follows,
\begin{equation}
m^* = \arg\underset{m }{{\min}}\  \mathcal{L}_{ME} \left (\theta,\phi \left ( m \right ) \right ) \label{mar}.
\end{equation}

\textbf{Bi-level learning strategy}. We adopt an online approximation strategy \cite{finn2017model,wang2022imbalanced} to jointly update $\{\theta ,\phi, m\}$ via the stochastic gradient decent (SGD) in an iterative manner. In each training iteration $t$, we sample a mini-batch of training data and move the current model parameters along the descent direction of MSL as follows,

\begin{small}
\begin{equation}
\left ( \hat{\theta} ^{t+1} ,\hat{\phi}^{t+1}\left ( m^t \right )   \right ) = \left ( \theta^{t} ,\phi^{t} \right )-\alpha \frac{1}{n}\sum_{i=1}^n\bigtriangledown_{\theta,\phi} \mathcal{L}_{MS}^i  \left ( \theta ,\phi, m \right )\bigg|_{\theta^t,\phi^t, m^t}, \label{6}
\end{equation}
\end{small}
where $n$ is batch size and $\alpha$ is the descent step size on $\{\theta ,\phi\}$. Then, a mini-batch of meta-validation data is sampled. We compute meta equalized loss on validation data using the updated model parameters $\{ \hat{\theta} ^{t+1} ,\hat{\phi}^{t+1}\left ( m^t \right ) \}$ and $m$ can be readily updated guided by,

\begin{footnotesize}
\begin{equation}
\begin{split}
m^{t+1}&=m^t-\beta \bigtriangledown_m\mathcal{L}_{ME}\left ( \theta,\phi \left ( m \right ) \right )\bigg|_{\hat{\theta }^{t+1},\hat{\phi }^{t+1},m^t} \\
&= m^t-\beta \frac{\partial  \phi(m)}{\partial m} \bigg|_{\hat{\phi }^{t+1},m^{t}} \frac{\partial \mathcal{L}_{ME}(\theta, \phi)}{\partial \phi } \bigg|_{\hat{\theta }^{t+1},\hat{\phi }^{t+1}}  \\
&= m^t+\frac{\alpha \beta }{n} \sum_{i=1}^{n}\frac{\partial^2 \mathcal{L}_{MS}^i \left ( \theta ,\phi, m \right )}{\partial \phi\partial m} \bigg|_{\theta^t,\phi^t,m^t} \frac{\partial \mathcal{L}_{ME}(\theta, \phi)}{\partial \phi } \bigg|_{\hat{\theta }^{t+1},\hat{\phi }^{t+1}} , \label{opt_margin}
\end{split}
\end{equation}
\end{footnotesize}
where $\beta$ is the descent step size on $m$. Note that the meta gradient update involves a second-order gradient, i.e., $\frac{\partial^2 \mathcal{L}_{MS}^i \left ( \theta ,\phi, m \right )}{\partial \phi\partial m}$, which can be easily implemented through popular deep learning frameworks like Pytorch. Since we make $m$ only directly related to $f_\phi ^{cls}$, we just need to unroll the gradient graph of $f_\phi ^{cls}$ and perform backward-on-backward automatic differentiation, which requires a lightweight overhead. Finally, the updated $m^{t+1}$ is employed to ameliorate $\{\theta,\phi\}$,

\begin{small}
\begin{equation}
\left ( \theta ^{t+1} ,\phi^{t+1} \right ) = \left ( \theta^{t} ,\phi^{t} \right )-\alpha \frac{1}{n}\sum_{i=1}^n\bigtriangledown_{\theta,\phi} \mathcal{L}_{MS}^i  \left ( \theta ,\phi, m \right )\bigg|_{\theta^t,\phi^t,m^{t+1}}. \label{opt_model}
\end{equation}
\end{small}
In summary, our MDN gradually ameliorates the model and margin parameters and ensures that 1) the learned model can be successfully debiased by minimizing the meta equalized loss over meta-validation data via Eq. (\ref{opt_margin}) and 2) it can simultaneously achieve satisfactory accuracy by minimizing the marginal softmax loss over training data via Eq. (\ref{opt_model}).

\section{Experiments}

This section evaluates the proposed MDN on the BiasedMNIST \cite{bahng2020learning}, Corrupted CIFAR-10 \cite{hendrycks2018benchmarking}, CelebA \cite{liu2015deep} and UTK-Face \cite{zhang2017age} datasets and demonstrates its effectiveness in debiasing with both quantitative and qualitative results.

\subsection{Datasets}

\textbf{BiasedMNIST} \cite{bahng2020learning} is a modified MNIST that contains 70K colored images of ten digits. The background color (bias attribute) is highly correlated with each digit category (target attribute). We set the correlation as $\rho \in  \{0.999, 0997, 0.995, 0.99\}$ to evaluate the effectiveness of our MDN under different imbalance ratios.

\textbf{Corrupted CIFAR-10} \cite{hendrycks2018benchmarking} is generated by corrupting the CIFAR-10 dataset designed for object classification. It consists of 60K images of 10 classes. We set the corruption as a bias attribute, and the object as a target attribute. We vary the ratio of bias-conflicting samples in training dataest and set it to be \{0.5\%, 1\%, 2\%, 5\%\} to simulate dataset bias during training.

\textbf{CelebA} \cite{liu2015deep} is a large-scale dataset for facial image recognition, containing more than 200K facial images annotated with 40 attributes. We set “male" and “young" as bias attributes, and select “attractive", “bags under eyes" and “big nose" as target attributes. 

\textbf{UTK-Face} \cite{zhang2017age} is a facial image dataset, which contains more than 20K images with annotations, such as age (young and old), gender (male and female) and race (white and others). We set “age" as a bias attribute, and select “gender" and “race" as target attributes. For constructing the biased training set, we truncate a portion of data to force the correlation between target and bias attributes to be $p(y|b) = 0.9$.

\begin{table*}
\centering
\begin{threeparttable}
\renewcommand\arraystretch{1.1}
\caption{Unbiased accuracy on BiasedMNIST dataset with various target-bias correlations. The best result is indicated in \textbf{\textcolor{red}{bold red}} and the second best result is indicated in \textcolor{blue}{\underline{undelined blue}}.}
 \label{BiasedMNIST}
        \setlength{\tabcolsep}{1.2mm}{
	\begin{tabular}{c|c|ccccccc|c}
		\toprule
         Corr & Vanilla & LNL \cite{kim2019learning} & ReBias \cite{bahng2020learning} & gDRO \cite{sagawa2019distributionally}& $\epsilon$-NCE \cite{barbano2022unbiased}  & LfF \cite{nam2020learning} & DI \cite{wang2020towards} & EnD \cite{tartaglione2021end} & \cellcolor{blue!5}\textbf{MDN (ours)}  \\ \hline \hline
         0.999 & 11.8\scriptsize $\pm$0.7 & 18.2\scriptsize$\pm$1.2 & 26.5\scriptsize $\pm$1.4 &  22.1\scriptsize$\pm$2.4 & 33.2\scriptsize$\pm$3.6 & 15.3\scriptsize$\pm$2.9 & 15.7\scriptsize$\pm$1.2 & \textcolor{blue}{\underline{59.5\scriptsize$\pm$2.3}} & \cellcolor{blue!5}\textbf{\textcolor{red}{74.3\scriptsize$\pm$1.8}} \\
         0.997 & 62.5\scriptsize $\pm$2.9 &   57.2\scriptsize $\pm$2.2 & 65.8\scriptsize $\pm$0.3 & 72.7\scriptsize $\pm$1.4 & 73.9\scriptsize $\pm$0.8 & 63.7\scriptsize $\pm$20.3 & 60.5\scriptsize $\pm$2.2 &  \textcolor{blue}{\underline{82.7\scriptsize $\pm$0.3}} & \cellcolor{blue!5}\textbf{\textcolor{red}{89.8\scriptsize $\pm$1.0}} \\
         0.995 & 79.5\scriptsize $\pm$0.1 &  72.5\scriptsize $\pm$0.9 & 75.4\scriptsize $\pm$1.0 & 83.3\scriptsize $\pm$0.2 & 83.7\scriptsize $\pm$0.4 & 90.3\scriptsize $\pm$1.4 & 89.8\scriptsize $\pm$2.0 &  \textcolor{blue}{\underline{94.0\scriptsize $\pm$0.6}} & \cellcolor{blue!5}\textbf{\textcolor{red}{94.1\scriptsize $\pm$0.5}} \\
         0.990 & 90.8\scriptsize $\pm$0.3 &  86.0\scriptsize $\pm$0.2 & 88.4\scriptsize $\pm$0.6 & 92.3\scriptsize $\pm$0.1 &  91.2\scriptsize $\pm$0.5 & 95.1\scriptsize $\pm$0.1 & \textbf{\textcolor{red}{96.9\scriptsize $\pm$0.1}} & 94.8\scriptsize $\pm$0.3 & \cellcolor{blue!5}\textcolor{blue}{\underline{95.9\scriptsize $\pm$0.1}} \\
         \bottomrule
         \end{tabular}}
    \begin{tablenotes}
    \scriptsize
        \item[*] $\epsilon$-NCE is an abbreviation of $\epsilon$-SupInfoNCE.
     \end{tablenotes}
\end{threeparttable}
\end{table*}

\begin{table*}
\renewcommand\arraystretch{1.1}
\caption{Unbiased accuracy on Corrupted CIFAR-10 dataset with varying ratio of bias-conflicting samples. The best result is indicated in \textbf{\textcolor{red}{bold red}} and the second best result is indicated in \textcolor{blue}{\underline{undelined blue}}.}
 \label{Corrupted CIFAR-10}
	\begin{center}
    \setlength{\tabcolsep}{1.5mm}{
	\begin{tabular}{c|c|ccccc|c}
		\toprule
         Ratio (\%) & Vanilla & HEX \cite{wang2019learning} & EnD \cite{tartaglione2021end}& ReBias \cite{bahng2020learning} & LfF \cite{nam2020learning} &  ECSGA \cite{zhao2023combating} &  \cellcolor{blue!5}\textbf{MDN (ours)}  \\ \hline \hline
         0.5 & 23.1\scriptsize$\pm$1.3 & 13.9\scriptsize $\pm$0.1 & 22.9\scriptsize $\pm$0.3 & 22.3\scriptsize $\pm$0.4 & 28.6\scriptsize $\pm$1.3 & \textbf{\textcolor{red}{30.9$\pm$0.8}} & \cellcolor{blue!5}\textcolor{blue}{\underline{30.3\scriptsize$\pm$0.9}}\\
         1.0 & 25.8\scriptsize $\pm$0.3 & 14.8\scriptsize $\pm$0.4 & 25.5\scriptsize $\pm$0.4 & 25.7\scriptsize $\pm$0.2 &33.1\scriptsize $\pm$0.8 &  \textcolor{blue}{\underline{36.2$\pm$0.4}} & \cellcolor{blue!5}\textbf{\textcolor{red}{36.3\scriptsize$\pm$0.6}} \\
         2.0 & 30.1\scriptsize $\pm$0.7 & 15.2\scriptsize $\pm$0.5 & 31.3\scriptsize $\pm$0.4 & 31.7\scriptsize $\pm$0.4 & 39.9\scriptsize $\pm$0.3  & \textcolor{blue}{\underline{44.1$\pm$1.2}} & \cellcolor{blue!5}\textbf{\textcolor{red}{45.2\scriptsize$\pm$0.6}} \\
         5.0 & 39.4\scriptsize $\pm$0.6 & 16.0\scriptsize $\pm$0.6 & 40.3\scriptsize $\pm$0.8 & 43.4\scriptsize $\pm$0.4 & 50.3\scriptsize $\pm$1.6 &\textcolor{blue}{\underline{55.1$\pm$1.2}} & \cellcolor{blue!5}\textbf{\textcolor{red}{57.8\scriptsize$\pm$1.1}} \\
         \bottomrule
         \end{tabular}}
    \end{center}
\end{table*}

\subsection{Implementation detail}

\textbf{Experimental setup}. All the experiments are implemented with PyTorch. We employ the four-layer CNN with kernel size 7$\times$7 for BiasedMNIST and ResNet-18 \cite{he2016deep} pretrained on ImageNet for the remaining datasets as our backbone architecture. We use the Adam optimizer with learning rates of $10^{-3}$ and $10^{-4}$ for \{Corrupted CIFAR-10, CelebA, UTK-Face\} and \{BiasedMNIST\}, respectively. The batch size is set to be 128 and 256 for \{BiasedMNIST, CelebA, UTK-Face\} and \{Corrupted CIFAR-10\}. We train the networks for 80, 250, 40, and 40 epochs for BiasedMNIST, Corrupted CIFAR-10, CelebA and UTK-Face, respectively, and the learning rates of margin parameters are $5\times 10^{-3}$, $3\times 10^{-2}$, $5\times 10^{-3}$ and $1\times 10^{-2}$. For CelebA, the size of the image is $224\times 224$, and we apply the augmentation of random flip. For UTK-Face, the size of the image is $64\times 64$, and we apply the augmentation of (1) random resized crop and (2) random flip. The balanced meta-validation set is dynamically re-sampling from the training set. 
We employ an online re-sampling strategy to ensure that the meta-validation set is continuously updated, which helps prevent the model from overfitting to it  to some extent.

\begin{table*}
\renewcommand\arraystretch{1.1}
\caption{Results of mitigating gender bias on CelebA dataset. The best result is indicated in \textbf{\textcolor{red}{bold red}} and the second best result is indicated in \textcolor{blue}{\underline{undelined blue}}. Here Y and B respectively represent target and bias attributes.}
 \label{celeba-gender}
	\begin{center}
    \footnotesize
    \setlength{\tabcolsep}{1.3mm}{
	\begin{tabular}{c|cccc|cccc|cccc|cccc}
		\toprule
         \multirow{3}*{Methods} &  \multicolumn{4}{c|}{Y=attractive, B=gender}  &  \multicolumn{4}{c|}{Y=big nose, B=gender} &  \multicolumn{4}{c|}{Y=bag under eye, B=gender} & \multicolumn{4}{c}{\cellcolor{gray!7}Avg.}\\
         & Unbias & Worst & Conflict &  EOD &  Unbias & Worst  & Conflict & EOD & Unbias &  Worst & Conflict & EOD  & \cellcolor{gray!7}Unbias & \cellcolor{gray!7}Worst & \cellcolor{gray!7}Conflict & \cellcolor{gray!7}EOD \\
         & ($\uparrow$) & ($\uparrow$) & ($\uparrow$) & ($\downarrow$)& ($\uparrow$) & ($\uparrow$) & ($\uparrow$) & ($\downarrow$)  & ($\uparrow$) & ($\uparrow$) & ($\uparrow$) & ($\downarrow$)  & \cellcolor{gray!7}($\uparrow$) & \cellcolor{gray!7}($\uparrow$) & \cellcolor{gray!7}($\uparrow$) & \cellcolor{gray!7}($\downarrow$) \\ \hline \hline
         Vanilla &  78.44 & 66.24 & 68.12 & 20.63 & 68.91 & 38.82 & 57.11 & 23.59 & 73.21 & 49.92 & 64.71 & 17.00  & \cellcolor{gray!7}73.52 & \cellcolor{gray!7}51.66 & \cellcolor{gray!7}63.31 & \cellcolor{gray!7}20.41  \\ \hline
         LfF \cite{nam2020learning}  & 77.42 & 67.46  & 67.56 & 19.70  & 70.80 & 42.45  & 59.87 & 21.85  & 72.91 & 50.55  & 65.43 & 14.95  & \cellcolor{gray!7}73.71 & \cellcolor{gray!7}53.49 & \cellcolor{gray!7}64.29 & \cellcolor{gray!7}18.83  \\
         LNL \cite{kim2019learning} & 78.37 & 63.21 & 67.49 & 21.75  & 69.75 & 32.18 & 53.65 & 32.20  & 76.55 & 64.22 & 66.40 & 20.29  & \cellcolor{gray!7}74.89 & \cellcolor{gray!7}53.20 & \cellcolor{gray!7}62.51 & \cellcolor{gray!7}24.75  \\
         DI \cite{wang2020towards}& 78.61 & 65.20 & 69.10 & 19.02 & 71.87 & 48.84 & 65.19 & 13.35 & 75.03 & 59.04 & 71.63 & \textcolor{blue}{\underline{6.78}}  &\cellcolor{gray!7}75.17 & \cellcolor{gray!7}57.69 & \cellcolor{gray!7}68.64 & \cellcolor{gray!7}13.05  \\
         EnD \cite{tartaglione2021end} & 78.83 & 67.72 & 68.76 & 20.15  & 70.71 & 38.59 & 61.19 & 19.03 & 73.46 & 52.21 & 74.93 & 8.36  & \cellcolor{gray!7}74.33 & \cellcolor{gray!7}52.84 & \cellcolor{gray!7}68.29 & \cellcolor{gray!7}15.85  \\
         ActiveSD \cite{zhang2022fair}  & 80.16 & \textcolor{blue}{\underline{73.98}} & 76.83 & 6.65  & 73.29 & 51.69 & 68.67 & 9.23 & \textcolor{blue}{\underline{77.93}} & \textbf{\textcolor{red}{70.17}} & 75.87 & \textbf{\textcolor{red}{6.03}} & \cellcolor{gray!7}77.13 & \cellcolor{gray!7}\textcolor{blue}{\underline{65.28}} & \cellcolor{gray!7}73.79 & \cellcolor{gray!7}7.30  \\
         BM \cite{qraitem2023bias} & \textbf{\textcolor{red}{80.59}} & 72.72 & \textcolor{blue}{\underline{77.37}} & \textcolor{blue}{\underline{6.44}} & \textcolor{blue}{\underline{74.59}} & \textcolor{blue}{\underline{62.68}} & \textbf{\textcolor{red}{75.01}} & \textbf{\textcolor{red}{3.21}} & 76.31 & 58.13 & \textcolor{blue}{\underline{78.38}} & 9.51 & \cellcolor{gray!7}\textcolor{blue}{\underline{77.16}} & \cellcolor{gray!7}64.51 & \cellcolor{gray!7}\textbf{\textcolor{red}{76.92}} & \cellcolor{gray!7}\textbf{\textcolor{red}{6.39}} \\ \hline
         \cellcolor{blue!5}\textbf{MDN (ours)}  & \cellcolor{blue!5}\textcolor{blue}{\underline{80.22}} & \cellcolor{blue!5}\textbf{\textcolor{red}{74.39}} & \cellcolor{blue!5}\textbf{\textcolor{red}{77.77}} & \cellcolor{blue!5}\textbf{\textcolor{red}{4.90}}  & \cellcolor{blue!5}\textbf{\textcolor{red}{75.09}} & \cellcolor{blue!5}\textbf{\textcolor{red}{70.18}} & \cellcolor{blue!5}\textcolor{blue}{\underline{71.24}} & \cellcolor{blue!5}\textcolor{blue}{\underline{7.70}} & \cellcolor{blue!5}\textbf{\textcolor{red}{78.72}} & \cellcolor{blue!5}\textcolor{blue}{\underline{69.49}} & \cellcolor{blue!5}\textbf{\textcolor{red}{78.42}} & \cellcolor{blue!5}8.32 & \cellcolor{blue!5}\textbf{\textcolor{red}{78.01}} & \cellcolor{blue!5}\textbf{\textcolor{red}{71.35}} & \cellcolor{blue!5}\textcolor{blue}{\underline{75.81}} & \cellcolor{blue!5}\textcolor{blue}{\underline{6.97}} \\
         \bottomrule
         \end{tabular}}
    \end{center}
\end{table*}

\textbf{Evaluation protocol}. To comprehensively evaluate the debiased performance of the proposed method, we consider four types of metrics, i.e., unbiased accuracy, worst-group accuracy, bias-conflict accuracy and equalized odds (EOD) \cite{mehrabi2021survey}. We divide the test set into different groups $g = (y, b)$, defined by a pair of target and bias attribute values. Unbiased accuracy denotes the average accuracy of different groups and worst-group accuracy is the worst accuracy among all groups. Bias-conflict accuracy reflects the model performance on bias-conflicting samples. EOD represents the parity of true positive rate (TPR) and false positive rate (FPR) between different bias classes. 
We explicitly construct a validation set and report the results at the epoch with the highest validation unbiased accuracy.

\begin{table*}
\renewcommand\arraystretch{1.1}
\caption{Results of mitigating age bias on CelebA dataset. The best result is indicated in \textbf{\textcolor{red}{bold red}} and the second best result is indicated in \textcolor{blue}{\underline{undelined blue}}. Here Y and B respectively represent target and bias attributes.}
 \label{celeba-age}
	\begin{center}
    \footnotesize
    \setlength{\tabcolsep}{1.3mm}{
	\begin{tabular}{c|cccc|cccc|cccc|cccc}
		\toprule
         \multirow{3}*{Methods} &  \multicolumn{4}{c|}{Y=attractive, B=age}  &  \multicolumn{4}{c|}{Y=big nose, B=age} & \multicolumn{4}{c|}{Y=bag under eye, B=age} &  \multicolumn{4}{c}{\cellcolor{gray!7}Avg.}\\
         & Unbias & Worst & Conflict &  EOD &  Unbias & Worst  & Conflict & EOD & Unbias &  Worst & Conflict & EOD  & \cellcolor{gray!7}Unbias & \cellcolor{gray!7}Worst & \cellcolor{gray!7}Conflict & \cellcolor{gray!7}EOD \\
         & ($\uparrow$) & ($\uparrow$) & ($\uparrow$) & ($\downarrow$)& ($\uparrow$) & ($\uparrow$) & ($\uparrow$) & ($\downarrow$)  & ($\uparrow$) & ($\uparrow$) & ($\uparrow$) & ($\downarrow$)  & \cellcolor{gray!7}($\uparrow$) & \cellcolor{gray!7}($\uparrow$) & \cellcolor{gray!7}($\uparrow$) & \cellcolor{gray!7}($\downarrow$) \\ \hline \hline
         Vanilla & 78.97 & 64.33 & 80.86 & 22.12  & 73.37 & 52.75 & 74.52 &  17.26  & 76.67 & 59.15 & 77.32 & 12.75 &  \cellcolor{gray!7}76.34 & \cellcolor{gray!7}58.74 & \cellcolor{gray!7}77.57 & \cellcolor{gray!7}17.38   \\ \hline
         LfF \cite{nam2020learning} &  77.93 & 66.36 & 79.32 & 22.10  & 73.64 & 53.38 & 75.23 & 15.80  & 76.32 & 59.15 & 76.83 & 10.05  & \cellcolor{gray!7}75.96  & \cellcolor{gray!7}59.63  & \cellcolor{gray!7}77.13 & \cellcolor{gray!7}15.98    \\
         LNL \cite{kim2019learning} & \textbf{\textcolor{red}{79.61}} & 63.28 & \textbf{\textcolor{red}{81.95}} & 22.53  & 74.11 & 52.16 & \textcolor{blue}{\underline{75.79}} & 15.44  & 76.70 & 59.76 & \textcolor{blue}{\underline{77.75}} & 14.30  & \cellcolor{gray!7}76.81 & \cellcolor{gray!7}58.40 & \cellcolor{gray!7}\textbf{\textcolor{red}{78.50}} & \cellcolor{gray!7}17.42   \\
         DI \cite{wang2020towards} & \textcolor{blue}{\underline{79.43}} & 65.58 & \textcolor{blue}{\underline{81.62}} & 17.32 & 74.37 & 59.28 & 74.26 & 13.11 & 77.23 & 65.66 & 76.39 & 4.63  & \cellcolor{gray!7}77.01 & \cellcolor{gray!7}63.51 & \cellcolor{gray!7}77.42 & \cellcolor{gray!7}11.69   \\
         EnD \cite{tartaglione2021end} & 78.22 & 63.81 & 78.38 & 21.40 & 75.04 & 57.17 & 72.68 & 10.14  & 76.38 & 62.53 & 75.84 & 5.76  & \cellcolor{gray!7}76.55 & \cellcolor{gray!7}61.17 & \cellcolor{gray!7}75.63 & \cellcolor{gray!7}12.43   \\
         ActiveSD \cite{zhang2022fair} & 78.47 & 63.93 & 78.12 & 17.65 & 75.73 & 62.65 & 75.72 & \textcolor{blue}{\underline{9.15}}  & 76.65 & 64.59 & 75.24 & \textbf{\textcolor{red}{3.95}}  & \cellcolor{gray!7}76.95 & \cellcolor{gray!7}63.72 & \cellcolor{gray!7}76.36 & \cellcolor{gray!7}10.25  \\
          BM \cite{qraitem2023bias} & 79.20 & \textcolor{blue}{\underline{67.52}} & 79.32 &  \textcolor{blue}{\underline{16.63}} & \textcolor{blue}{\underline{76.02}} & \textcolor{blue}{\underline{67.87}} &  74.74 & \textbf{\textcolor{red}{5.17}} & \textcolor{blue}{\underline{77.74}} & \textcolor{blue}{\underline{66.34}} & 75.19 & 5.09 & \cellcolor{gray!7}\textcolor{blue}{\underline{77.65}} & \cellcolor{gray!7}\textcolor{blue}{\underline{67.24}} & \cellcolor{gray!7}76.42 & \cellcolor{gray!7}\textbf{\textcolor{red}{8.96}} \\ \hline
         \cellcolor{blue!5}\textbf{MDN (ours)}   & \cellcolor{blue!5}78.95 & \cellcolor{blue!5}\textbf{\textcolor{red}{68.79}} & \cellcolor{blue!5}79.44 & \cellcolor{blue!5}\textbf{\textcolor{red}{15.10}} & \cellcolor{blue!5}\textbf{\textcolor{red}{76.95}} & \cellcolor{blue!5}\textbf{\textcolor{red}{69.22}} & \cellcolor{blue!5}\textbf{\textcolor{red}{76.03}} & \cellcolor{blue!5}9.20 & \cellcolor{blue!5}\textbf{\textcolor{red}{79.38}} & \cellcolor{blue!5}\textbf{\textcolor{red}{74.79}} & \cellcolor{blue!5}\textbf{\textcolor{red}{77.89}} & \cellcolor{blue!5}\textcolor{blue}{\underline{4.45}} & \cellcolor{blue!5}\textbf{\textcolor{red}{78.43}} & \cellcolor{blue!5}\textbf{\textcolor{red}{70.93}} & \cellcolor{blue!5}\textcolor{blue}{\underline{77.79}} & \cellcolor{blue!5}\textcolor{blue}{\underline{9.58}} \\
         \bottomrule
         \end{tabular}}
    \end{center}
\end{table*}

\subsection{Synthetic experiments}

\textbf{Results on BiasedMNIST}. Table \ref{BiasedMNIST} reports the unbiased accuracy scores on BiasedMNIST with different bias ratios for the training set. We can observe that the vanilla method is indeed susceptible to dataset bias, leading to diminished performance on unbiased test sets. As the spurious correlation between target classes and bias attributes increases, the model becomes more inclined to favor over-represented samples and show lower unbiased accuracy. For example, the performance of the vanilla method is 90.8\% when it is trained on the dataset with $\rho=0.99$ and gets down to 11.8\% when trained on the dataset with $\rho=0.999$. Our approach consistently outperforms the vanilla model and obtains unbiased accuracies of 74.3\%, 89.8\%, 94.1\% and 95.9\% under different bias ratios. Notably, it is even superior to group DRO \cite{sagawa2019distributionally} and EnD \cite{tartaglione2021end} in most cases, especially when the spurious correlation is higher in the training set. These results show evidence that our approach is less affected by the background color.

\textbf{Results on Corrupted CIFAR-10}. We also evaluate our model on Corrupted CIFAR-10 and show the unbiased accuracy results in Table \ref{Corrupted CIFAR-10}. Due to the challenging scenario, existing methods exhibit only marginal improvements in unbiased accuracy on Corrupted CIFAR-10. For example, ReBias \cite{bahng2020learning} promotes Hilbert-Schmidt independence between the network prediction and all biased predictions, but it only increases the unbiased accuracy from 39.4\% to 43.4\% at 5\% ratio of bias-conflicting samples.
ECSGA \cite{zhao2023combating} performs the second best in terms of fairness, which reweights bias-aligned and bias-conflicting samples according to their currently produced gradient contributions.
Again, our proposed method significantly improves the debiasing performances regardless of the bias severities. Compared with ECSGA \cite{zhao2023combating}, MDN obtains an average gain of 0.8\% across different ratios, proving the effectiveness of the added adaptive margins.

\begin{table*}
\renewcommand\arraystretch{1.1}
\caption{Results of mitigating age bias on UTK-Face dataset. The best result is indicated in \textbf{\textcolor{red}{bold red}} and the second best result is indicated in \textcolor{blue}{\underline{undelined blue}}. Here Y and B respectively represent target and bias attributes.}
 \label{utk}
	\begin{center}
    \footnotesize
    \setlength{\tabcolsep}{1.5mm}{
	\begin{tabular}{c|cccc|cccc|cccc}
		\toprule
         \multirow{3}*{Methods} &  \multicolumn{4}{c|}{Y=gender, B=age}  &  \multicolumn{4}{c|}{Y=race, B=age}
         & \multicolumn{4}{c}{\cellcolor{gray!7}Avg.}\\
         & Unbias & Worst & Conflict &  EOD &  Unbias &  Worst & Conflict & EOD & \cellcolor{gray!7}Unbias & \cellcolor{gray!7}Worst  & \cellcolor{gray!7}Conflict & \cellcolor{gray!7}EOD \\
          & ($\uparrow$) & ($\uparrow$) & ($\uparrow$) & ($\downarrow$)  & ($\uparrow$) & ($\uparrow$) & ($\uparrow$) & ($\downarrow$) & \cellcolor{gray!7}($\uparrow$) & \cellcolor{gray!7}($\uparrow$) & \cellcolor{gray!7}($\uparrow$) & \cellcolor{gray!7}($\downarrow$) \\ \hline \hline
         Vanilla  & 72.19 & 6.54 & 45.73 & 52.91  & 79.05 & 57.73 & 61.86 & 34.36  & \cellcolor{gray!7}75.62 & \cellcolor{gray!7}32.14 & \cellcolor{gray!7}53.80 &  \cellcolor{gray!7}43.64    \\ \hline
         LfF \cite{nam2020learning}  & 72.95 & 7.14  & 46.84 & 52.22 & 79.38 & 62.99  & 62.22 & 34.33  & \cellcolor{gray!7}76.17 & \cellcolor{gray!7}35.07 & \cellcolor{gray!7}54.53 &  \cellcolor{gray!7}43.28   \\
         LNL \cite{kim2019learning} & 71.20 & 5.35 & 44.16 & 54.08  & 79.67 & 57.83 & 61.91 &  35.50  & \cellcolor{gray!7}75.44 & \cellcolor{gray!7}31.59 & \cellcolor{gray!7}53.04 & \cellcolor{gray!7}44.79   \\
         RWG \cite{idrissi2022simple} & 74.03 & 16.66 & 53.38 & 41.30 & 85.32 & 73.00 & 77.74 & 15.16  & \cellcolor{gray!7}79.68 & \cellcolor{gray!7}44.83 & \cellcolor{gray!7}65.56 &  \cellcolor{gray!7}28.23    \\
         DI \cite{wang2020towards} & 74.86 & 23.80 & 57.01 & 36.83  & \textbf{\textcolor{red}{87.59}} & \textcolor{blue}{\underline{83.99}} & \textcolor{blue}{\underline{84.21}} & 6.75  & \cellcolor{gray!7}81.23 & \cellcolor{gray!7}53.90 & \cellcolor{gray!7}70.61 & \cellcolor{gray!7}21.79    \\
         EnD \cite{tartaglione2021end} & 72.39 & 4.16 & 46.61 & 51.55  & 77.62 & 54.12 & 59.56 & 36.12  & \cellcolor{gray!7}75.01 & \cellcolor{gray!7}29.14 & \cellcolor{gray!7}53.09 & \cellcolor{gray!7}43.84    \\
         ActiveSD \cite{zhang2022fair}  & 76.99 & 38.69 & 64.45 & 32.87  & 87.25 & 82.99 & \textbf{\textcolor{red}{85.10}} & \textcolor{blue}{\underline{4.30}} & \cellcolor{gray!7}\textcolor{blue}{\underline{82.12}} & \cellcolor{gray!7}60.84 & \cellcolor{gray!7}74.78 & \cellcolor{gray!7}\textcolor{blue}{\underline{18.59}}   \\ 
         BM \cite{qraitem2023bias} & \textbf{\textcolor{red}{78.70}} & \textcolor{blue}{\underline{56.55}} & \textcolor{blue}{\underline{72.52}} & \textcolor{blue}{\underline{28.73}} & \textcolor{blue}{\underline{87.29}} & 79.24 & 82.12 & 10.33 & \cellcolor{gray!7}\textbf{\textcolor{red}{83.00}} & \cellcolor{gray!7}\textcolor{blue}{\underline{67.90}} & \cellcolor{gray!7}\textcolor{blue}{\underline{77.32}} & \cellcolor{gray!7}19.53 \\ \hline
         \cellcolor{blue!5}\textbf{MDN (ours)}   & \cellcolor{blue!5}\textcolor{blue}{\underline{78.33}} & \cellcolor{blue!5}\textbf{\textcolor{red}{61.21}} &  \cellcolor{blue!5}\textbf{\textcolor{red}{76.79}} & \cellcolor{blue!5}\textbf{\textcolor{red}{24.02}}  & \cellcolor{blue!5}85.62 & \cellcolor{blue!5}\textbf{\textcolor{red}{83.99}} & \cellcolor{blue!5}84.03 & \cellcolor{blue!5}\textbf{\textcolor{red}{3.17}} & \cellcolor{blue!5}81.98 & \cellcolor{blue!5}\textbf{\textcolor{red}{72.60}} & \cellcolor{blue!5}\textbf{\textcolor{red}{80.41}} & \cellcolor{blue!5}\textbf{\textcolor{red}{13.60}}  \\
         \bottomrule
         \end{tabular}}
    \end{center}
\end{table*}

\subsection{Real-world experiments}

\textbf{Results on CelebA}. Tables \ref{celeba-gender} and \ref{celeba-age} report the results on CelebA dataset to evaluate the effectiveness of our method in mitigating biases with respect to gender and age, respectively. From the results, we have the following observations. First, the vanilla method cannot perform fairly across different samples, which proves the negative effect of spurious correlations. For example, most of the “attractive" images in the training dataset are of women. As a result, the vanilla model achieves 89.31\% on “attractive" women but only obtains 66.24\% on “attractive" men (worst-group accuracy). Second, existing debiased methods can improve the model fairness to some extent. DI \cite{wang2020towards} trains separated target classifiers for each bias class, so as to make the target prediction independent of the bias attribute.
ActiveSD \cite{zhang2022fair}, which learns controllable shortcut features for each sample to eliminate bias information from target features, performs best in terms of the EOD metric when using “bag under eye" as the target attribute. Finally, our proposed MDN outperforms the comparison methods on most tasks. It achieves 78.01\%, 71.35\% and 6.97 in terms of unbiased accuracy, worst-group accuracy and EOD when mitigating gender bias, and simultaneously obtains 78.43\%, 70.93\% and 9.58 when mitigating age bias.

\textbf{Results on UTK-Face}. Table \ref{utk} shows the debiased performance on UTK-Face dataset. The observations are similar to the ones on CelebA. 1) The vanilla method without fairness constraint usually exhibits low values of fairness metrics. 2) Some previous debiased methods, e.g., ActiveSD \cite{zhang2022fair} and DI \cite{wang2020towards}, improve model fairness compared with the vanilla method; while others perform poorly, e.g., LNL \cite{kim2019learning} and EnD \cite{tartaglione2021end}. We conjecture that this is because more serious dataset bias of UTK-Face makes it difficult for some simple methods to disentangle bias information from target prediction. 3) Our MDN addresses the bias problem from the perspective of margin. During training, we introduce margin penalty and reshape the decision boundary rather than removing bias information to achieve fairness. 
From the results, we can see that our MDN obtains better debiased performance, especially it improves the worst-group accuracy from 38.69\% to 61.21\% compared with ActiveSD when mitigating age bias for gender prediction. Compared with the re-sampling methods, MDN outperforms RWG \cite{idrissi2022simple} which re-weights the sampling probability of each example to achieve group-balance. Additionally, it successfully reduces EOD from 19.53 to 13.60 compared with BM \cite{qraitem2023bias} which is a class-conditioned sampling method. This demonstrates the advantage of the modified margin objective and the effectiveness of considering intra-class variation for bias-conflicting data. 

\begin{table*}
\renewcommand\arraystretch{1.1}
\caption{Ablation study on UTK-Face dataset. The best result is indicated in \textbf{bold}. Here Y and B respectively represent target and bias attributes.}
 \label{ablation}
	\begin{center}
    \footnotesize
    \setlength{\tabcolsep}{1.5mm}{
	\begin{tabular}{l|cc|cccc|cccc|cccc}
		\toprule
         \multirow{3}*{Methods} &  \multirow{2}*{MSL} & \multirow{2}*{MEL} & \multicolumn{4}{c|}{Y=gender, B=age}  &  \multicolumn{4}{c|}{Y=race, B=age}
         & \multicolumn{4}{c}{\cellcolor{gray!7}Avg.}\\
         &  &  & Unbias & Worst & Conflict &  EOD &  Unbias &  Worst & Conflict & EOD & \cellcolor{gray!7}Unbias & \cellcolor{gray!7}Worst  & \cellcolor{gray!7}Conflict & \cellcolor{gray!7}EOD \\
         &  &  & ($\uparrow$) & ($\uparrow$) & ($\uparrow$) & ($\downarrow$)  & ($\uparrow$) & ($\uparrow$) & ($\uparrow$) & ($\downarrow$) & \cellcolor{gray!7}($\uparrow$) & \cellcolor{gray!7}($\uparrow$) & \cellcolor{gray!7}($\uparrow$) & \cellcolor{gray!7}($\downarrow$) \\ \hline \hline
         Vanilla & \textcolor{red}{\ding{55}} & \textcolor{red}{\ding{55}}  & 72.19 & 6.54 & 45.73 & 52.91  & 79.05 & 57.73 & 61.86 & 34.36  & \cellcolor{gray!7}75.62 & \cellcolor{gray!7}32.14 & \cellcolor{gray!7}53.80 &  \cellcolor{gray!7}43.64    \\
         Vanilla+Re-sample& \textcolor{red}{\ding{55}} & \textcolor{red}{\ding{55}}  & 74.03 & 16.66 & 53.38 & 41.30 & 85.32 & 73.00 & 77.74 & 15.16  & \cellcolor{gray!7}79.68 & \cellcolor{gray!7}44.83 & \cellcolor{gray!7}65.56 &  \cellcolor{gray!7}28.23    \\ \hline
         MDN w/o MSL& \textcolor{red}{\ding{55}} & \textcolor{green}{\ding{51}}  & 50.00 & 0.00 & 50.00 & 0.00  & 50.00 & 0.00 & 50.00 & 0.00  & \cellcolor{gray!7}50.00 & \cellcolor{gray!7}0.00 & \cellcolor{gray!7}50.00 & \cellcolor{gray!7}0.00    \\
         MDN w/o MEL & \textcolor{green}{\ding{51}} & \textcolor{red}{\ding{55}}  & 73.43 & 21.42 & 55.01 & 36.84 & 84.84 & 70.99 & 77.99 &  13.69  & \cellcolor{gray!7}79.14 & \cellcolor{gray!7}46.21 & \cellcolor{gray!7}66.50 & \cellcolor{gray!7}25.27    \\ \hline
         \cellcolor{blue!5}\textbf{MDN (ours)}  & \cellcolor{blue!5}\textcolor{green}{\ding{51}} & \cellcolor{blue!5}\textcolor{green}{\ding{51}}  & \cellcolor{blue!5}\textbf{78.33} & \cellcolor{blue!5}\textbf{61.21} & \cellcolor{blue!5}\textbf{76.79} & \cellcolor{blue!5}\textbf{24.02}  & \cellcolor{blue!5}\textbf{85.62} & \cellcolor{blue!5}\textbf{83.99} & \cellcolor{blue!5}\textbf{84.03} & \cellcolor{blue!5}\textbf{3.17}  & \cellcolor{blue!5}\textbf{81.98} & \cellcolor{blue!5}\textbf{72.60} & \cellcolor{blue!5}\textbf{80.41} & \cellcolor{blue!5}\textbf{13.60}  \\
         \bottomrule
         \end{tabular}}
    \end{center}
\end{table*}

\subsection{Ablation study}

We report the ablation analysis of MSL and MEL in our MDN via extensive variant experiments on UTK-Face dataset. The results are shown in Table \ref{ablation}.

\textbf{Effectiveness of MSL}. We remove MSL from MDN and denote it as \emph{MDN w/o MSL}. Since the meta learning framework is utilized to learn the optimal margin for MSL, it is also removed as well. Therefore, \emph{MDN w/o MSL} directly uses cross-entropy loss and MEL to optimize the model on training set. From the results, we can find that it easily suffers from the collapse of network. The reason may be as follows. For each target class, MEL is designed similarly to EOD metric which aims to reduce the performance skewness between different bias classes. The model is prone to find a shortcut to achieve this goal, for example, the accuracies on old men, young men, old women and young women are 0\%, 0\%, 100\% and 100\% respectively when mitigating age bias for gender prediction. Instead of directly optimizing the model parameters by MEL, our MDN introduces a margin penalty into cross-entropy loss and utilizes MEL to optimize the margin parameters to trade off bias-conflicting and bias-aligned samples, which can avoid the wrong shortcut.

\begin{figure*}
\centering
\subfigure[Vanilla: Y=“big nose"]{
\label{visual_ce_bignose} 
\includegraphics[height=3cm]{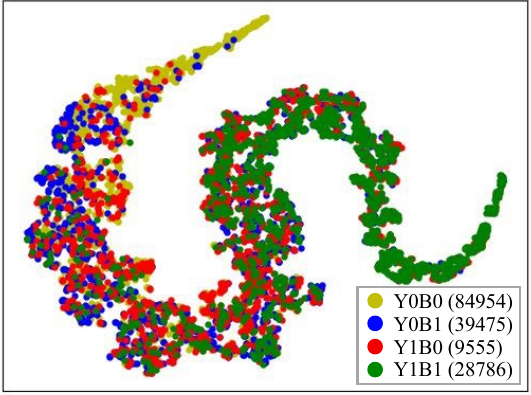}}
\hspace{0cm}
\subfigure[MDN: Y=“big nose"]{
\label{visual_mar_bignose} 
\includegraphics[height=3cm]{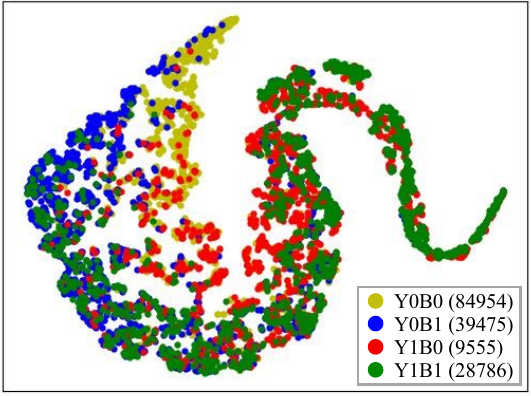}}
\hspace{0cm}
\subfigure[Vanilla: Y=“bags under eyes"]{
\label{visual_ce_bageye} 
\includegraphics[height=3cm]{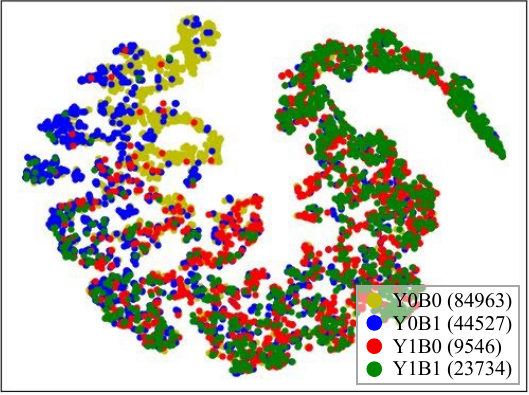}}
\hspace{0cm}
\subfigure[MDN: Y=“bags under eyes"]{
\label{visual_mar_bageye} 
\includegraphics[height=3cm]{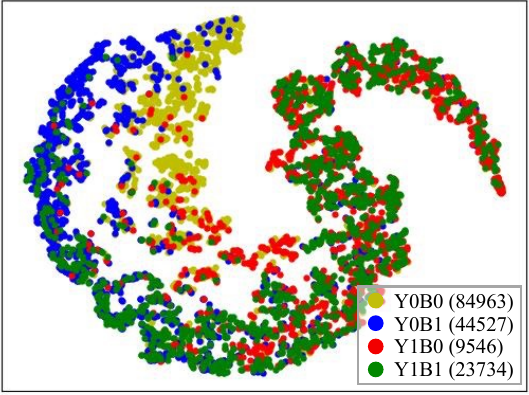}}
\caption{t-SNE \cite{maaten2008visualizing} embedding visualizations on CelebA when mitigating gender bias. Here Y and B respectively represent target and bias attributes. \emph{Y1B0} denotes the group of the images with target label $y=1$ and bias label $b=0$, and similarly for \emph{Y0B0}, \emph{Y0B1} and \emph{Y1B1}, respectively. The number of training images in each group is shown in the bracket.}
\label{visual} 
\end{figure*}

\textbf{Effectiveness of MEL}. We also perform an ablation study with \emph{MDN w/o MEL} in which the margin parameters are updated via meta-optimization guided by the unbiased loss, rather than by MEL. The definition of unbiased loss is similar to that of unbiased accuracy, and it denotes the average (cross-entropy) loss of different groups. Compared with the vanilla method, \emph{MDN w/o MEL} can improve the performance of the worst group, and thus achieve fairness. However, it is still inferior to MDN which proves the effectiveness of our proposed MEL.

\textbf{Comparison with re-sampling}. Since the meta-validation set in our MDN is constructed by dynamically re-sampling from the training set, we further compare MDN with the re-sampling method. As seen from the results in Table \ref{ablation}, MDN improves the worst-group accuracy from 44.83\% to 72.60\% and decreases EOD from 28.23 to 13.60 on average compared with the re-sampling method. Although the re-sampling method can also shift the focus from bias-aligned samples to bias-conflicting ones, it fails to address the problem of intra-class variation for bias-conflicting samples and results in limited improvement of generalization capability. Different from it, MDN can reshape the decision boundary by margin penalty and strongly squeeze their intra-class variations to improve generalization.

\begin{table*}
\renewcommand\arraystretch{1.1}
\caption{Conformal inference on BiasedMNIST dataset with various target-bias correlations. The best result is indicated in \textbf{\textcolor{red}{bold red}} and the second best result is indicated in \textcolor{blue}{\underline{undelined blue}}.}
 \label{conformal}
	\begin{center}
    \footnotesize
    \setlength{\tabcolsep}{1.8mm}{
	\begin{tabular}{l|cccc|cccc|cccc|cccc}
		\toprule
         \multirow{3}*{Methods} &  \multicolumn{4}{c|}{Corr=0.999}  &  \multicolumn{4}{c|}{Corr=0.997} &  \multicolumn{4}{c|}{Corr=0.995} & \multicolumn{4}{c}{Corr=0.990}\\
         &  Cvg & Size & CvgD &  SizeD &  Cvg & Size & CvgD &  SizeD & Cvg & Size & CvgD &  SizeD  & Cvg & Size & CvgD &  SizeD \\
         & ($\uparrow$) & ($\downarrow$) & ($\downarrow$) & ($\downarrow$)& ($\uparrow$) & ($\downarrow$) & ($\downarrow$) & ($\downarrow$) & ($\uparrow$) & ($\downarrow$) & ($\downarrow$) & ($\downarrow$) & ($\uparrow$) & ($\downarrow$) & ($\downarrow$) & ($\downarrow$) \\ \hline \hline
         Vanilla &  0.889 & 7.845 & 0.111 & 2.842 & 0.903 & 2.900 & 0.097 & 1.753 & 0.924 & 1.475 & 0.076 & 0.627  & 0.959 & 1.322 & 0.041 & 0.317  \\ \hline
         gDRO \cite{sagawa2019distributionally} & 0.888 & 6.810 & 0.112 & 4.534  & \textcolor{blue}{\underline{0.916}} & \textcolor{blue}{\underline{2.322}} & 0.084 & 1.385  & 0.931 & 1.931 & 0.069 & 0.708  & 0.972 & 1.241 & 0.028 & 0.363  \\
         DI \cite{wang2020towards} & \textcolor{blue}{\underline{0.894}} & 8.375 & \textcolor{blue}{\underline{0.106}} & \textbf{\textcolor{red}{0.864}} & 0.893 & 2.696 & \textcolor{blue}{\underline{0.082}} & \textcolor{blue}{\underline{0.642}} & \textcolor{blue}{\underline{0.947}} & \textbf{\textcolor{red}{1.071}} & \textbf{\textcolor{red}{0.003}} & \textbf{\textcolor{red}{0.023}}  & \textcolor{blue}{\underline{0.978}} & \textbf{\textcolor{red}{1.018}} & \textbf{\textcolor{red}{0.001}} & \textbf{\textcolor{red}{0.001}}  \\
         EnD \cite{tartaglione2021end} & 0.860 & \textcolor{blue}{\underline{5.613}} & 0.132 & \textcolor{blue}{\underline{0.956}}  & 0.883 & 2.931 & 0.117 & 1.778 & 0.929 & 2.161 & 0.071 & 1.243  & 0.956 & 1.484 & 0.044 & 0.521   \\ \hline
         \cellcolor{blue!5}\textbf{MDN(ours)}  & \cellcolor{blue!5}\textbf{\textcolor{red}{0.914}} & \cellcolor{blue!5}\textbf{\textcolor{red}{2.138}} & \cellcolor{blue!5}\textbf{\textcolor{red}{0.080}} & \cellcolor{blue!5}1.067  & \cellcolor{blue!5}\textbf{\textcolor{red}{0.961}} & \cellcolor{blue!5}\textbf{\textcolor{red}{1.379}} & \cellcolor{blue!5}\textbf{\textcolor{red}{0.030}} & \cellcolor{blue!5}\textbf{\textcolor{red}{0.416}} & \cellcolor{blue!5}\textbf{\textcolor{red}{0.984}} & \cellcolor{blue!5}\textcolor{blue}{\underline{1.204}} & \cellcolor{blue!5}\textcolor{blue}{\underline{0.014}} & \cellcolor{blue!5}\textcolor{blue}{\underline{0.217}} & \cellcolor{blue!5}\textbf{\textcolor{red}{0.986}} & \cellcolor{blue!5}\textcolor{blue}{\underline{1.224}}  & \cellcolor{blue!5}\textcolor{blue}{\underline{0.014}} & \cellcolor{blue!5}\textcolor{blue}{\underline{0.189}} \\
         \bottomrule
         \end{tabular}}
    \end{center}
\end{table*}

\subsection{Analysis}

\textbf{Conformal inference disparity.} Conformal prediction \cite{angelopoulos2020uncertainty} is a method of distribution-free uncertainty quantification, which constructs a set of predictions that provably covers the true diagnosis with a high probability. Formally, the uncertainty set function $C(X)$ guarantees states: $P(Y \in C(X)) \ge 1-\alpha$, where $X$ and $Y$ are the unseen test example and the true class respectively, and $\alpha$ is a pre-specified confidence level. We conduct conformal inference using RAPS \cite{angelopoulos2020uncertainty} and calculate \emph{coverage (Cvg)} and \emph{set size (Size)} on BiasedMNIST dataset to prove the effectiveness of our method. Following \cite{lu2022fair}, we also show \emph{coverage disparity (CvgD)} and \emph{set size disparity (SizeD)} to quantify bias. \emph{CvgD} and \emph{SizeD} are the differences in \emph{Cvg} and \emph{Size} between bias-aligned and bias-conflicting samples. Table \ref{conformal} shows the results on BiasedMNIST dataset, where $\alpha$ is set to be 0.1. As seen from the results, our MDN can obtain satisfactory performance under different target-bias correlations. \emph{CvgD} and \emph{SizeD} are successfully reduced to improve fairness, especially when the spurious correlation is higher.

\begin{figure*}
\centering
\subfigure[Y=“big nose"]{
\label{margin_bignose} 
\includegraphics[height=3.8cm]{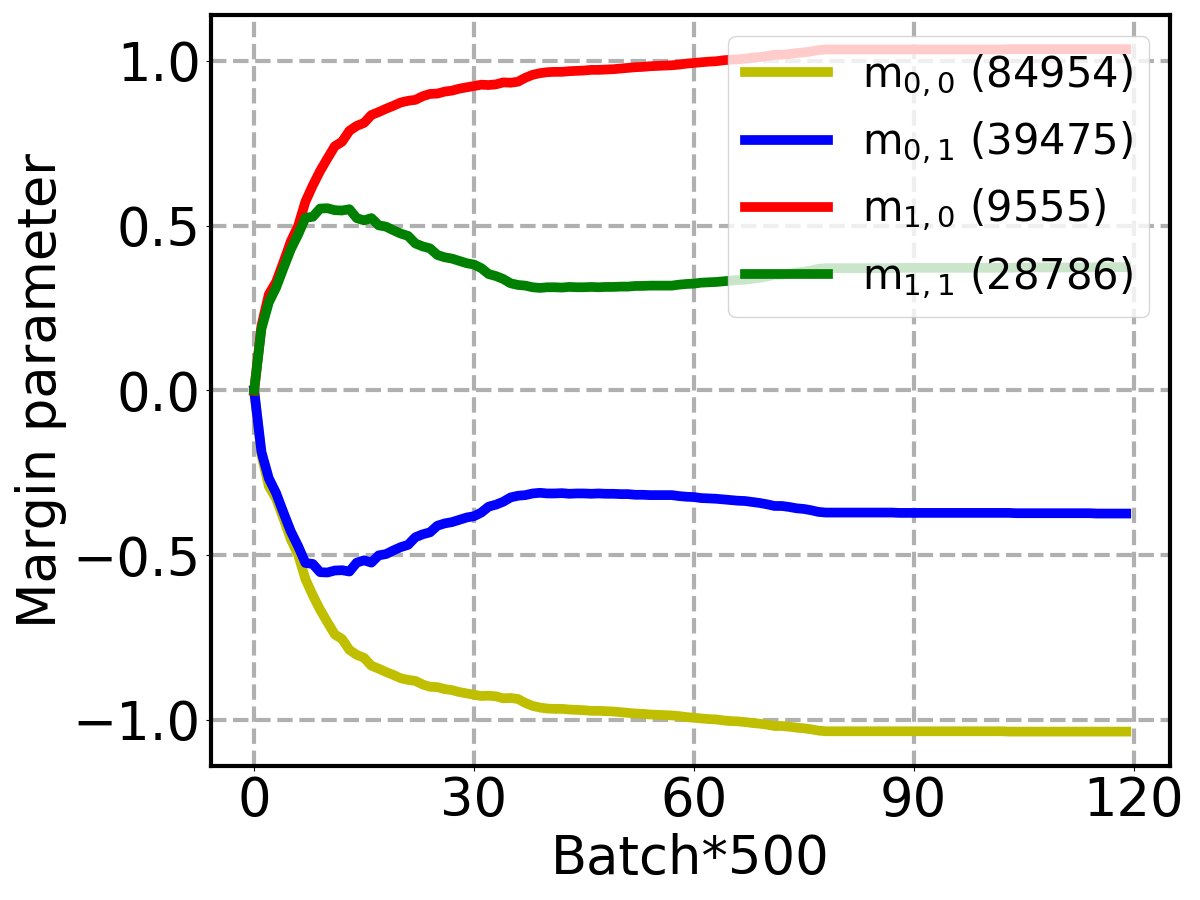}}
\hspace{1cm}
\subfigure[Y=“bags under eyes"]{
\label{margin_bagseyes} 
\includegraphics[height=3.8cm]{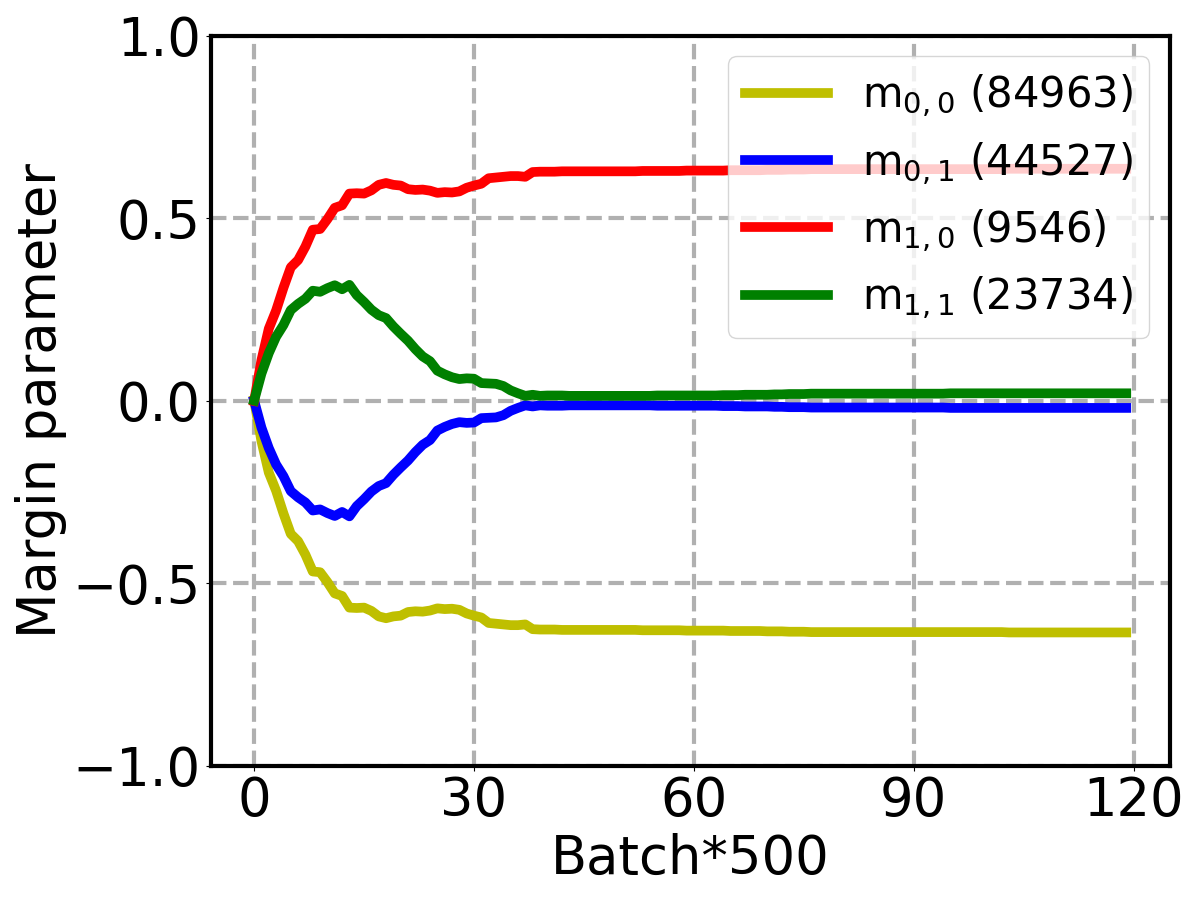}}
\hspace{1cm}
\subfigure[Y=“attractive"]{
\label{margin_attravtive} 
\includegraphics[height=3.8cm]{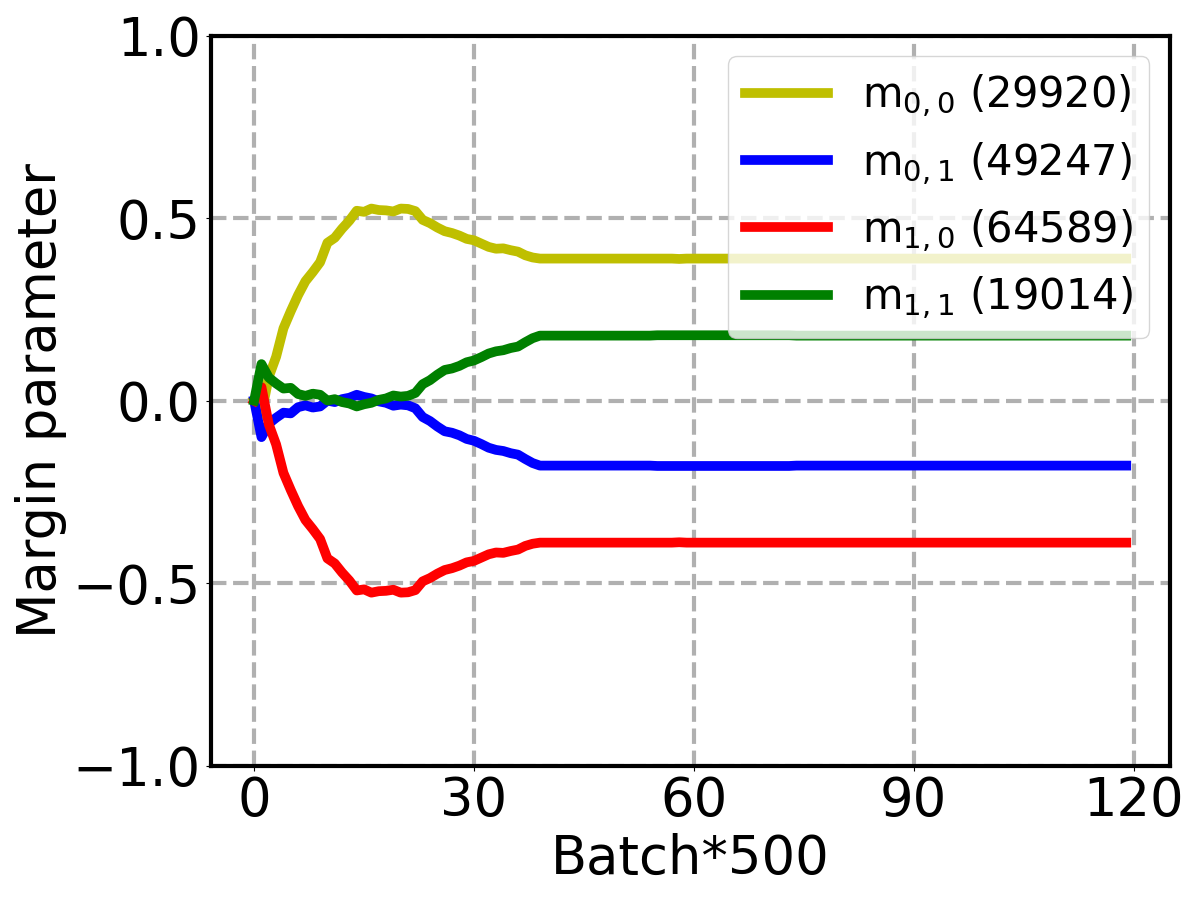}}
\caption{The margin parameters learned for different groups on CelebA when mitigating gender bias. Here Y and B respectively represent target and bias attributes. $m_{1,0}$ denotes the margin parameter learned for the images with target label $y=1$ and bias label $b=0$, and similarly for $m_{0,0}$, $m_{0,1}$ and $m_{1,1}$, respectively. The number of training images in each group is shown in the bracket.}
\label{visual_margin} 
\end{figure*}

\textbf{Feature visualization}. Fig. \ref{visual} shows the t-SNE \cite{maaten2008visualizing} visualization of the features extracted by the vanilla method and MDN on CelebA test set. The features are divided into 4 groups (i.e., \emph{Y0B0}, \emph{Y0B1}, \emph{Y1B0} and \emph{Y1B1}) in terms of target class and bias attribute, which are visualized in different colors. \emph{Y1B0} denotes the group of the images with target label $y=1$ and bias label $b=0$, and similarly for \emph{Y0B0}, \emph{Y0B1} and \emph{Y1B1}, respectively. From the results in Fig. \ref{visual_ce_bignose} and \ref{visual_ce_bageye}, we can observe that the features belonging to different target classes fail to be separated well in the vanilla method and the degree of class separation varies across different groups. Especially, compared with other groups, \emph{Y1B0} (visualized by red points) is more prone to mix up with other target classes (visualized by blue and yellow points) since \emph{Y1B0} is under-represented in training data. This skewness across different groups shows a serious bias problem. As shown in Fig. \ref{visual_mar_bignose} and \ref{visual_mar_bageye}, our MDN successfully pushes the red points away from the blue and yellow ones, which improves the model performance on bias-conflicting samples and thus achieves fairness. This phenomenon proves that our proposed method can make the model emphasize the minority groups and squeeze their intra-class variations by introducing a margin penalty.

\textbf{Margin visualization}. To better understand our MDN, we further plot the learned margin parameters of different groups during the training process in Fig. \ref{visual_margin}. As we describe above, the training set is separated into 4 groups in terms of target class and bias attribute. First, we can observe that $m$ stably converges with the increased number of iteration. Second, the value of margin parameter is always inversely proportional to the image number. For example, as shown in Fig. \ref{margin_bignose}, our method automatically learns the largest margin parameter $m_{1,0}$ for \emph{Y1B0} since this group is infrequent. This is consistent with our assumption that we prefer stricter constraints for bias-conflicting samples, making the model pay more attention to them and squeeze their intra-class variations. Third, MDN can adaptively adjust the margin parameters according to the dataset skewness. As shown in Fig. \ref{margin_bagseyes} and \ref{margin_attravtive}, the data bias is more serious when selecting “bags under eyes” as the target attribute than when selecting “attractive". Accordingly, the skewness of the margin parameters across different groups is increased to better trade off bias-conflicting and bias-aligned samples when the training dataset is more biased. Benefiting from this adaptive learning mechanism, our method can successfully mitigate bias under various imbalance ratios.

\begin{table}
\renewcommand\arraystretch{1.1}
\caption{Training cost analysis on BiasedMNIST dataset.}
 \label{cost}
	\begin{center}
    \setlength{\tabcolsep}{2.8mm}{
	\begin{tabular}{l|rrr}
		\toprule
         Methods & Params & GPU memory	& Training time \\ \hline
         Vanilla & 0.53 M & 1439 MiB & \textasciitilde 58 min \\
         EnD \cite{tartaglione2021end} &  0.53 M  & 1437 MiB  & \textasciitilde 123 min \\ 
         DI \cite{wang2020towards} & 0.54 M & 1439 MiB & \textasciitilde 59 min    \\
         LfF \cite{nam2020learning} & 1.06 M & 1641 MiB & \textasciitilde 149 min    \\ 
         MDN (ours) & 0.53 M & 1639 MiB & \textasciitilde 97 min     \\
         \bottomrule
         \end{tabular}}
    \end{center}
\end{table}

\textbf{Training cost analysis.} 
Table \ref{cost} shows the training cost analysis on BiasedMNIST dataset. Our experiments on BiasedMNIST dataset are implemented in Python on a desktop with one Tesla T4 GPU and Intel Xeon Gold 5218 CPU of 2.3GHz. During training phase, MDN includes both inner loop training and outer loop optimization. The gradient calculations and optimization steps within the inner loop may lead to increased computational overhead compared with the vanilla method. To enhance efficiency, we make $m$ directly only related to the linear classifier rather than the whole classification network, which requires a lightweight overhead. As seen from the results, the increase in computational cost of our proposed MDN is acceptable compared to the baseline model, especially considering the significant performance improvement of MDN. Moreover, the training cost of our MDN is comparable to other existing debiased methods such as LfF and EnD.

\section{Conclusion}

In this work, we proposed a novel marginal debiased network (MDN) for debiasing in visual recognition. We introduced a margin penalty to move the model focus from bias-aligned samples to bias-conflicting ones, preventing the model from learning the unintended correlations. The decision boundary in training was reshaped by margins, which can help the model accommodate the distribution shift between training and testing and thus improve the generalization. Additionally, we proposed to adaptively learn the optimal margin parameters by meta learning to better trade off bias-conflicting and bias-aligned samples. Experiments on various synthetic and real-world datasets including BiasedMNIST, Corrupted CIFAR-10, CelebA and UTKFace showed that our method successfully outperformed the previous debiased approaches.

\textbf{Limitations and future work}. First, our method requires full knowledge of the bias labels which are unavailable in most practical scenarios. Our further research will delve into investigating a more relaxed condition of MDN that does not require bias labels. Second, to ensure a fair comparison with other methods without introducing additional data, the balanced meta-validation set is constructed by dynamically re-sampling from the training set in MDN. How to get a better validation set to guide the learning of margin parameters and avoid overfitting remains to be explored. Third, when the number of bias-conflicting samples in the training set is extremely small, MDN may fail. In the future, we plan to combine our method with generative models.

{
\bibliographystyle{IEEEtran}
\bibliography{egbib}
}

\end{document}